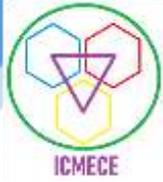

# Reinforcement learning based local path planning for mobile robot

Mehmet Gök
*Kahramanmaraş İstiklal University*
Kahramanmaraş, Turkey
mehmet.gok@istiklal.edu.tr
ORCID: 0000-0003-1656-5770

Mehmet Tekerek
*Kahramanmaraş Sütçüimam University*
Kahramanmaraş, Turkey
tekerek@ksu.edu.tr
ORCID: 0000-0001-6112-3651

Hamza Aydemir
*Kahramanmaraş Sütçüimam University*
Kahramanmaraş, Turkey
aydemirhamza1402@gmail.com
ORCID: 0000-0002-2657-3195

*Abstract*—Different methods are used for a mobile robot to go to a specific target location. These methods work in different ways for online and offline scenarios. In the offline scenario, an environment map is created once, and fixed path planning is made on this map to reach the target. Path planning algorithms such as A* and RRT (Rapidly-Exploring Random Tree) are the examples of offline methods. The most obvious situation here is the need to re-plan the path for changing conditions of the loaded map. On the other hand, in the online scenario, the robot moves dynamically to a given target without using a map by using the perceived data coming from the sensors. Approaches such as SFM (Social Force Model) are used in online systems. However, these methods suffer from the requirement of a lot of dynamic sensing data.

Thus, it can be said that the need for re-planning and mapping in offline systems and various system design requirements in online systems are the subjects that focus on autonomous mobile robot research. Recently, deep neural network powered Q-Learning methods are used as an emerging solution to the aforementioned problems in mobile robot navigation.

In this study, machine learning algorithms with deep Q-Learning (DQN) and Deep DQN architectures, are evaluated for the solution of the problems presented above to realize path planning of an autonomous mobile robot to avoid obstacles.

*Keywords—mobile robot, reinforcement learning, deep q-learning, path planning*

## I. INTRODUCTION

Path planning algorithms, which have been the subject of research since the beginning of robotic technology, aim to provide a collision-free path from the starting position to the desired target position. The quality of the planned path is key to the robot's ability to complete the task accurately and efficiently [1]. In the field of autonomous mobile robots' path planning, there are two basic scenarios, online and offline [2]. Offline scenario, an environment map is created once, and fixed path planning is made on this map to reach the target. Path planning algorithms such as A* and RRT (Rapidly-Exploring Random Tree) are examples of offline methods [3]. On the other hand, in the online scenario, the robot moves dynamically to a given target without using a map by using the perceived data coming from the sensors. Artificial potential field method [4], fuzzy control [5], neural network [6] and reinforcement learning [2] methods are widely used for online path planning in autonomous mobile robots [7]. Among these methods, the reinforcement learning approach, in which only sparse information is given to the learning agent in the form of a scalar reward signal beside the current sensory information, is a good solution for path planning [2].

Path planning algorithms can be run with the Robot Operating System (ROS). ROS is an open-source meta-operating system licensed under the BSD license (Open-Source Initiative) that enables efficient development of robot systems. ROS includes; packages developed for various robot platforms, libraries designed for robot applications, navigation packages and 3D robot simulation environments [8]. Integrated or independently developed path planning algorithms can be simulated and tested in ROS robot simulation environments. Through the simulation environment, new concepts, strategies and algorithms can be developed with cost and time savings. The ROS-based Gazebo robot simulation environment is a powerful tool for testing robotic applications such as path planning [9].

Reinforcement learning is an effective machine learning method that learns by feedback from the environment interaction. In this context, reinforcement learning is a framework for obtaining a behavioral policy that maximizes value through trial and error, even under unknown conditions. Deep learning is a method of high-level pattern recognition [10].

Reinforcement learning has been tested in various studies for the task of path planning. A trained model using Q-Learning is proposed to establish a path for grid-based decomposition of the environment from point A to point B [4]. Q-Learning was applied for industrial robot motion planning [11]. DDPG (Deep Deterministic Policy Gradient) reinforcement learning algorithm was used to plan the movement of bipedal robots in a football match [12]. It has been proposed to use reinforcement learning for trajectory planning in autonomous vehicles [13]. A path for efficient exploration in unknown environments was created through the Actor-Critic Reinforcement learning model [14]. The GRU RNN (Recurrent Neural Network) was used to plan a path from point A to point B while avoiding obstacles in a grid-based environment [15]. Also, different approaches [7, 16] that use DQN and DDQN architectures for navigation has been proposed.

In this study, machine learning algorithms with deep Q-Learning (DQN) and Deep DQN architectures, are evaluated for the solution of the problems presented above to realize path planning of an autonomous mobile robot to avoid obstacles. Current implementation of DQN approach provided with Turtlebot3 Machine Learning approach is enhanced.

## II. REINFORCEMENT LEARNING

Reinforcement Learning (RL) is an effective machine learning method that learns by feedback from the environment interaction. The agent is placed in an unknown environment, takes an action and observes the state of the environment for each action. Also, after an Agent executes an action, it will get feedback from the environment. This is a feedback mechanism that employs a "trial and error" process. While a good action choice results in a positive reward; a bad action choice results in a negative reward. According to this process, mapping a state to an action is called a policy or a strategy. The reinforcement learning model is shown in Fig. 1.

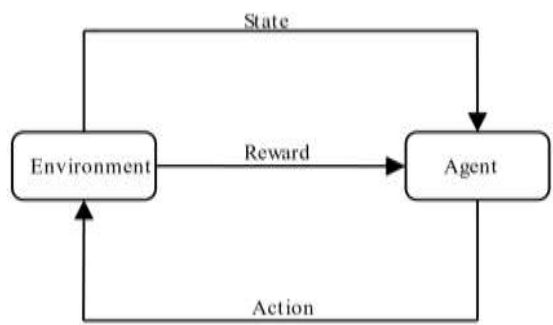

**Figure 1**. The process of reinforcement learning.

## III. DEEP Q-LEARNING ALGORITHM (DQN)

Reinforcement learning can be implemented with very common algorithms such as Q-learning. However, traditional Q-Learning algorithms require a Q-table that is not sufficient for high-dimensional problem spaces. Thus, a deep neural network is used instead of Q-Table and the policy is determined via neural network outputs.

Q learning is an off-policy RL algorithm that is based on the Bellman Equation. The main aim is to discover the policy that maximizes the total reward. At each training iteration, Q values are updated in Q-Table and at the end of the algorithm, the best policy that provides the solution is obtained. Q-Table update process is given as in Eq. (1).

$$Q(s,a) \leftarrow Q(s,a) + \alpha \left[ r(s,a) + \gamma \max_{a'} Q(s',a') - Q(s,a) \right] \quad (1)$$

where $s$ is the state or observation, $a$ is the action, $r$ is the reward, $\alpha$ is the learning rate, $\gamma$ is the discount factor. The discount factor helps in making the cumulative sum of rewards bounded over time, so more immediate rewards have higher values [16].

To solve dimensionality problems, DQN has been developed with the combined use of DNN and RL algorithm. The discount factor causes rewards to lose their value over time, Deep Neural Network (DNN) is used to obtain appropriate Q values in DQN. For the training phase, the agent's experiences are stored and sampled randomly from the experience replay memory. It has been proven that the use of experience replay is successful in providing good stability. Experience replay buffer includes action and state observations at each step as shown in Eq. (2).

$$e_t = (s_t, a_t, r_t, s_{t+1}) \quad (2)$$

where $e_t$ is the experience stored at time $t$, $s_t$ is the state at time $t$, $a_t$ is the action at time $t$, $r_t$ is the reward at time $t$, $s_{t+1}$ is the state at time $t+1$ [7, 16].

Q-Learning and DQN block diagrams are given in Fig. 2.

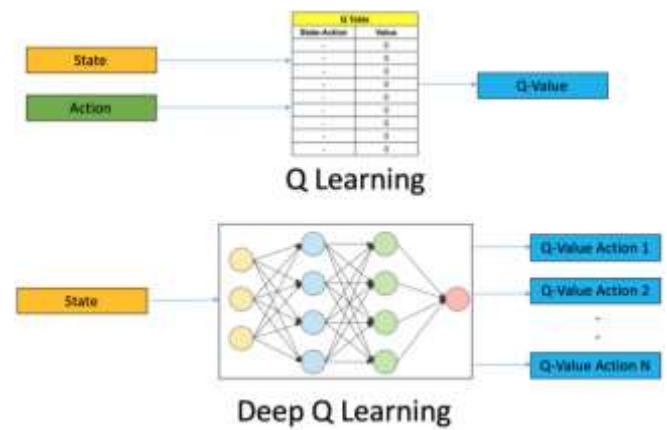

**Figure 2**. Use of multilayer perceptron instead of Q-Table.

In this study, we enhance the Double DQN structure for TurtleBot3 to train a better local path planner. D-DQN uses two networks; online and target networks. This approach is proposed by [17] to solve the overestimation problem. Online network is used to select the action and the target network is used to evaluate the action values [3]. The current solution provided by ROBOTIS e-Manual [18] uses dual networks, but does not implement a fully D-DQN approach. Reward function is executed by using the traditional DQN algorithm. It uses action values of online network. We use action values from target network.

In mobile robot studies, a DQN agent learns an optimal policy to navigate the robot from point a to point b with minimum effort. Fig. 3. depicts simple reinforcement learning with DQN:

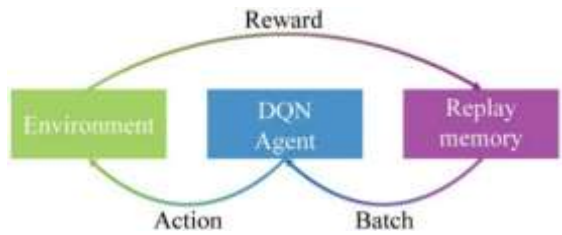

**Figure 3.** Reinforcement learning approach.

- The agent selects the action by Q-value and reflects it in the environment

- It stores the reward obtained in an environment in replay memory.

- The agent is updated by randomly sampling the stored samples from the replay memory.

- The agent's goal is to get the TurtleBot3 to a target state.

D-DQN approach is implemented with Python2, Keras and Google TensorFlow library. ROS Melodic, Gazebo and Ubuntu 18.04 Linux consist the infrastructure of the application. TurtleBot3 simulation pack is opted to use DDQN structure. For each episode of the training, TurtleBot3 gets a reward from the Gazebo environment. When the time limit is exceeded (episode step is selected as 500) or TurtleBot3 model crashes to an obstacle; trained agent gets a negative reward. Each action turns the robot heading towards the target, bringing a positive reward for the agent. The episode simulation is reset when the robot reaches the target, crashes into an obstacle or time limit situation.

## IV. SETTING PARAMETERS

The task of the DQN Agent in this application is to get the TurtleBot3 robot to its destination by avoiding obstacles. TurtleBot3 receives a positive reward when it approaches the target after the selected action, and a negative reward (punishment) when it gets farther from the target. Each training stage ends when TurtleBot3 crashes into an obstacle or ends after a certain period of time. This can be treated as a after time-step limit exceeds. During the episode, when TurtleBot3 reaches the target, it gets a very large positive reward. When it hits an obstacle, it gets a very large negative reward.

### A. State Set

State is the observation of the environment in which the agent is located and the definition of the current situation. In this application, the size of the state space (state_size) is 26. First 24 elements of the state are the distance information coming from Laser Distance Sensor (LDS). Last two are; distance to the goal, and the heading angle to the goal (Fig. 4).

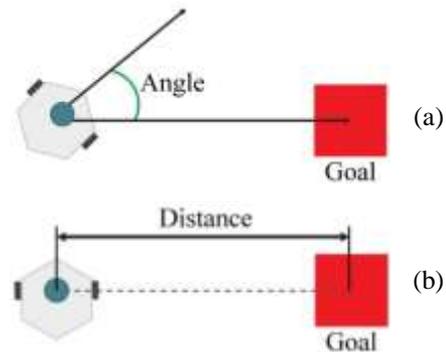

**Figure 4.** Heading angle to the goal (a) and distance from goal (b) inputs.

The 2D distance information incoming from LDS of Turtlebot3 presents 360 samples by default. The number of samples is decreased to 24 to reduce the neural network input size (state size).

### B. Set Action

Action can be defined as the action chosen by the agent according to the situation. In this application, TurtleBot3 has a constant linear velocity of 0.15 m/s, and the angular velocity is determined by the selected action. There are five actions which are related to different directions (Fig. 5). These actions compensate the heading angle to the goal for Turtlebot3 by changing angular velocity. The velocity values are listed in Table I.

TABLE III. ACTION SET DEFINITONS

| Action | Angular Velocity (rad/s) |
|---|---|
| 0 | -1.5 |
| 1 | -0.75 |
| 2 | 0 |
| 3 | 0.75 |
| 4 | 1.5 |

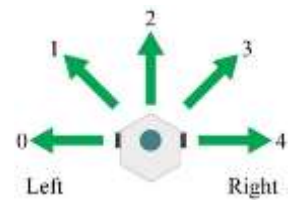

**Figure 5.** Angular directions related to actions.

### C. Set Reward

When TurtleBot3 takes an action in any state, it receives a reward. Determining the reward value for the learning action is of critical importance. A reward can be negative or positive. The *setReward* function can change and be updated according to the designer. The reward function executes when the robot interacts with an obstacle or reaches a goal by providing feedback.

In this application, the reward function is designed in two stages. In the first step, an $R_\theta$ (reward from angle) value is obtained in the angular direction. Then, an $R_d$ (reward from distance) value is obtained from the distance to the target. Obtaining these values are shown in Fig. 6, Fig. 7, and Fig. 8.

$$\theta = \frac{\pi}{2} + action * \frac{\pi}{8} + \phi$$

$$R_\theta = 5 * 1^{-\theta}$$

ϕ: Yaw of TurtleBot3

θ: Angle from Goal

$R_\theta$: Reward from angle

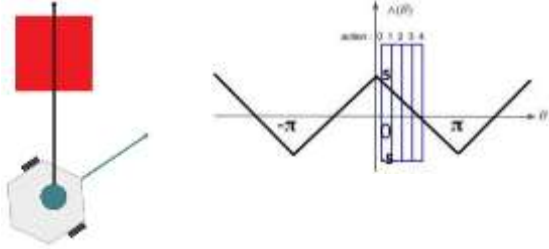

**Figure 6.** Angular direction calculation.

$$R_d = 2^{\frac{D_c}{D_g}}$$

$D_c$: Current distance from Goal

$D_g$: Absolute distance from Goal

$R_d$: Reward from distance

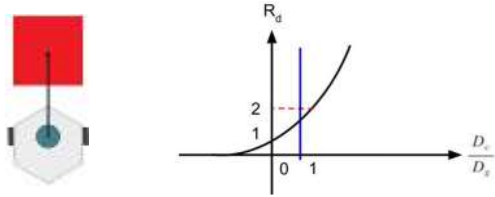

**Figure 7.** Linear direction calculation.

if $-\frac{1}{2}\pi < \theta < \frac{1}{2}\pi$, $R_\theta \geq 0$

else $R_\theta < 0$

if $D_c < D_g$, $R_d > 2$

else $1 < R_d \leq 2$

$R = R_d * R_\theta$

θ : Angle from TurtleBot3 to Goal

$D_c$: Current distance from Goal

$D_g$ : Absolute distance from Goal

$R_\theta$: Reward from θ

$R_d$: Reward from distance

$R$: Reward

**Figure 8.** Reward function calculation

Calculations related to the reward function are used as in ROBOTIS TurtleBot3 e-Manual. In the gazebo simulation, the robot goes towards the randomly selected target and performs the algorithm with this reward function. Each time the robot reaches the target, an episode is completed and the reward-action-status-next-state is saved in the buffer reserved for re-experiencing (experience replay buffer). At each stage of the training, a random group is selected from this buffer and the training is performed.

### D. Setting the Hyperparameters

In the gazebo simulation, the robot goes towards the randomly selected target and performs the algorithm with this reward function. Each time the robot reaches the target, an episode is completed and the reward-action-status-next-state is saved in the buffer reserved for re-experiencing. At each stage of the training, a random group is selected from this buffer and the training is performed. TurtleBot3 machine learning hyperparameters are shared in the table.

TABLE IV. HYPERPARAMETERS FOR DDQN ALGORITHM

| Hyperparameter | Default value | Definition |
|---|---|---|
| episode_step | 6000 | The time step of one episode. |
| target_update | 2000 | Update rate of target network. |
| discount_factor | 0.99 | Represents how much future events lose their value according to how far away. |
| learning_rate | 0.00025 | Learning speed. If the value is too large, learning does not work well, and if it is too small, learning time is long. |
| epsilon | 1.0 | The probability of choosing a random action. |
| epsilon_decay | 0.99 | Reduction rate of epsilon. When one episode ends, the epsilon reduces. |
| epsilon_min | 0.05 | The minimum of epsilon. |
| batch_size | 64 | Size of a group of training samples. |
| train_start | 64 | Start training if the replay memory size is greater than 64. |
| memory | 1000000 | The size of replay memory. |

In this method, the robot is used as an agent and the motion model is trained in a complex environment. The trained model is loaded on the robot and it is ensured that it arrives at the desired destination in the online working environment.

### V. EXPERIMENTAL RESULTS

Total reward graph related to the current DDQN solution is depicted in Fig. 9. during the training for 2000 episodes. Total reward graph related to the proposed DDQN solution is shown in Fig. 10. According to the graphs, proposed solution presents more stable learning with higher average scores than current DDQN solution.

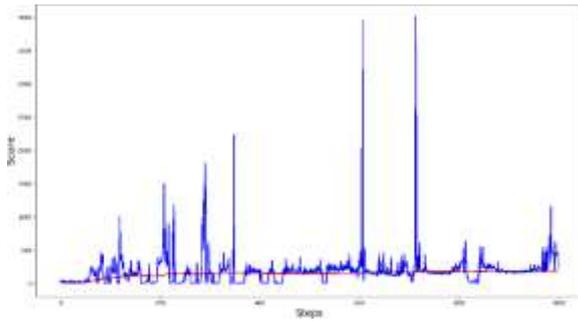

**Figure 9.** Total reward graph related to current DDQN solution.

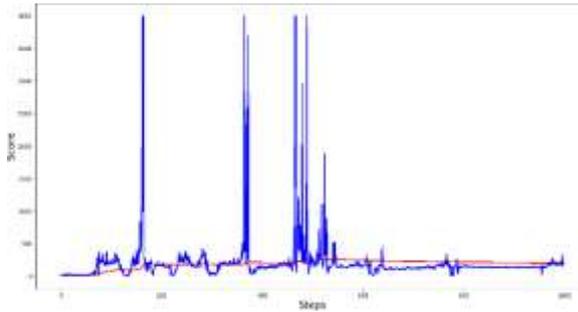

**Figure 10.** Total reward graph related to current DDQN solution.

## VI. CONCLUSION

In this study, we enhance the Double DQN structure for Turtlebot3 provided by ROBOTIS by using target network action values. According to the training and inference results, DDQN approaches can be used as alternative solution for better local path planning for mobile robots.

ACKNOWLEDGMENT

This study conducted in KSU Robotics Laboratory
.